\documentclass[11pt]{article}

\usepackage[final]{acl}

\usepackage{times}
\usepackage{latexsym}
\usepackage[T1]{fontenc}
\usepackage[utf8]{inputenc}
\usepackage{inconsolata}
\usepackage[normalem]{ulem}
\usepackage{multirow}
\usepackage{rotating}
\usepackage{enumitem}
\usepackage{verbatim}
\usepackage{listings}
\usepackage{caption}
\usepackage{placeins}

\usepackage{amsmath}
\usepackage{amssymb}
\usepackage{mathtools}
\usepackage{amsthm}

\usepackage{booktabs} 

\usepackage{microtype}

\usepackage{inconsolata}

\usepackage{graphicx}

\graphicspath{{../}{./}}

\title{EnComp: Lightweight Encoder-Only Context Compression for Retrieval-Augmented Question Answering}

\author{
  \textbf{Thao Do}, \textbf{Dinh Phu Tran}, \textbf{An Vo}, \textbf{Seon Kwon Kim}, \textbf{Daeyoung Kim} \\
  School of Computing, KAIST, Daejeon, Republic of Korea \\
  \texttt{\{thaodo, phutx2000, an.vo, lukaskim, kimd\}@kaist.ac.kr}
}

\begin{document}
\maketitle
\begin{abstract}
Efficient context compression is critical for retrieval-augmented question answering in resource-constrained settings, where long retrieved contexts increase latency, memory use, and LLM reader cost. We propose a lightweight encoder-only framework for query-driven sentence pruning that preserves answer-critical evidence while aggressively reducing irrelevant context. Our method learns marginal contribution scores for sentences using counterfactual training signals and optimizes a contrastive ranking objective that separates critical evidence from noncritical context. Our approach scores all sentences from a single full-context encoding, enabling fast inference with low computational overhead. Experiments show that it maintains accuracy comparable to the strongest baseline while using 3.7$\times$ less peak memory and achieving nearly 3$\times$ lower compression latency, demonstrating an effective quality--efficiency trade-off for practical resource-constrained deployment.
\end{abstract}

\section{Introduction}
Retrieval-augmented generation (RAG) has emerged as a powerful paradigm for enhancing
large language models (LLMs) with external knowledge, improving factual accuracy and
reducing hallucinations \cite{lewis2020retrieval,ram2023context}. However, retrieving
more documents improves coverage yet expands context length, increasing memory
consumption and inference latency \cite{lu-etal-2025-turborag}, while also introducing
extraneous content that can distract the model and degrade generation quality
\cite{liu2023lost}.

\begin{figure}[t]
    \centering
    \includegraphics[width=\linewidth]{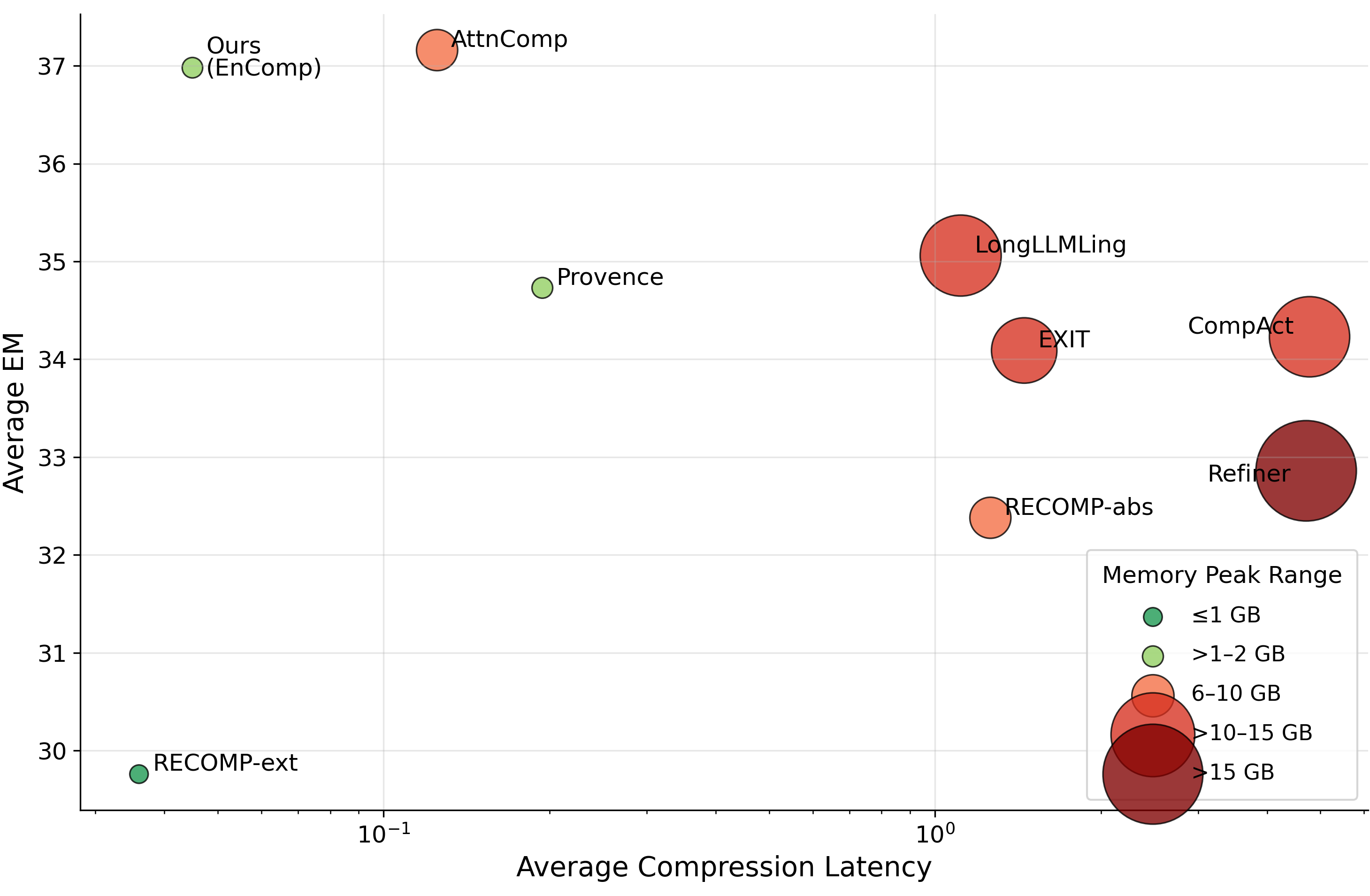}
    \caption{Quality--latency--memory trade-off across five datasets using Llama-3.3-70B-Instruct as the reader at retrieval depth $k=20$. EM denotes average exact-match answering accuracy, while bubble size indicates the peak VRAM required. Our method achieves competitive EM while requiring 3.7$\times$ less VRAM and nearly 3$\times$ lower compression latency than the strongest baseline.}
    \label{fig:tradeoff}
\end{figure}

Context compression has emerged as a potential solution to this challenge, enabling RAG systems to retain essential information while reducing computation \cite{li2023compressing,wingate2022prompt}. Existing approaches broadly fall into two categories: abstractive methods that generate condensed summaries~\cite{yoon2024compact,xu2024recomp,li2024refiner} and extractive methods that select relevant text segments~\cite{jiang2023longllmlingua}. Abstractive methods achieve high compactness by rewriting content into concise summaries, but their token-by-token generation process incurs additional latency, which can exceed the time saved from reduced context length. Extractive methods generally offer faster compression because they avoid generation, but many rely on rigid selection criteria that fail to adapt to varying query requirements or preserve evidence distributed across multiple sentences.

Recent approaches have sought to mitigate some of these limitations. EXIT~\cite{hwang2024exit} and AttnComp~\cite{luo-etal-2025-attncomp} use context-aware extractive compression that leverages full-document context to make sentence-level or chunk-level decisions in parallel, enabling more efficient filtering over long contexts. However, they still rely on billion-parameter LLMs for high filtering quality, creating computational overhead and high memory requirements for a sentence selection task. Meanwhile, comparative studies show that encoder-only models remain competitive on various natural language understanding (NLU) tasks~\cite{benayas2025comparative,nielsen2024encoder}. In a related direction, Provence~\cite{chirkova2025provence} pursues efficient pruning with token-level supervision. However, using sentence relevance labels as supervision for every token, including non-informative words, can introduce noise since many tokens in an important sentence (e.g., function words) are not themselves informative. Its sentence rounding heuristic further makes sentence selection depend on aggregated token predictions rather than direct sentence-level scoring.

In this work, we revisit the fundamental design choices in query-driven context pruning to address the trade-off among quality, latency, and memory requirements. Our key premise is that compression should not treat each sentence as an isolated relevance label, but should estimate how much the sentence contributes to the answerability of the full passage. We therefore propose a lightweight encoder-only compressor that learns sentence-level contribution scores under full-passage context. During training, we estimate these contributions by counterfactually removing sentences and measure the resulting change in passage answerability. These counterfactual redactions provide soft supervision for these contributions, while a contrastive ranking objective encourages answer-critical evidence to outrank distracting context. At inference time, the model scores all sentences from a single full-context encoding, keeping compression simple, fast, and memory-efficient. Our main contributions are as follows:

\begin{enumerate}[itemsep=1pt]
    \item We introduce \textbf{EnComp}, a lightweight encoder-only framework for query-driven context compression. Its main novelty is to model sentence selection as marginal contribution scoring for passage answerability where EnComp learns these scores from counterfactual redaction signals and contrastive ranking, while requiring only a single full-context encoding at inference time.
    
    \item We demonstrate the effectiveness and efficiency of EnComp through comprehensive experiments on five standard QA benchmarks with both open-source and proprietary LLM readers. Compared with recent strong baselines, EnComp preserves competitive answer fidelity and compression quality while substantially reducing compression latency and GPU memory usage.
\end{enumerate}

\section{Related Work}

\subsection{Retrieval-Augmented Generation}

Retrieval-Augmented Generation (RAG) \cite{lewis2020retrieval, gao2023retrieval} improves LLMs by grounding generation in external knowledge sources. Recent iterative, multi-hop, and recursive retrieval methods \cite{khattab2022demonstrate, trivedi2022interleaving} refine queries or retrieve additional documents during generation to access dispersed evidence, but often increase latency and token usage while introducing irrelevant details that can reduce precision \cite{liu2023lost}. This has motivated token-reduction and document-compression techniques that preserve essential evidence while supporting efficient, high-quality generation.

\subsection{Query-aware Context Compression in RAG}

Context compression in RAG can be divided into soft and hard compression. Soft compression shortens embedding-level representations \cite{mu2023learning, ge2023context}, offering controllable rates but often requiring fine-tuning or architectural changes. Notably, COCOM~\cite{rau2025context} shows that query-independent soft compression can be effective and efficient when context embeddings for a fixed corpus are precomputed offline and the reader is adapted to consume them. Hard compression filters or rewrites natural-language text. Abstractive methods summarize retrieved passages with generative models \cite{xu2024recomp, yoon2024compact, li2024refiner}, achieving high token reduction but often increasing resource use and latency.

Extractive methods, such as RECOMP-Extr \cite{xu2024recomp} and the LLMLingua family \cite{jiang2023longllmlingua, pan2024llmlingua}, select salient sentences or tokens from retrieved documents, preserving original evidence and reducing hallucination risk. However, query-agnostic selection can disrupt coherence or omit critical information. LongLLMLingua adds query awareness and configurable compression rates, but balancing coverage and efficiency remains challenging. EXIT \cite{hwang2024exit} prunes context under the full-context condition, yet its architecture introduces high compression cost, and binary thresholding yields modest compression gains. AttnComp~\cite{luo-etal-2025-attncomp} uses a billion-parameter decoder-only LLM as an encoder and gates chunks with cumulative attention weights during prefill, but still incurs memory overhead and potentially produces unstable compression rates due to coarse chunk granularity.

Meanwhile, Provence~\cite{chirkova2025provence} improves efficiency through token-level pruning, but token-centric supervision is only an indirect proxy for sentence-level salience. Aggregating token predictions with heuristics may miss discourse- and structure-level cues needed for coherent, utility-driven extraction. 

To balance quality, latency, and memory efficiency in hard compression, we adopt extractive sentence selection with a lightweight encoder-only architecture. Our method directly estimates each sentence's clue contribution under full-passage context, preserving answer-critical evidence while keeping compression fast and memory-light. EnComp targets online query-dependent compression, orthogonal to offline query-independent soft compression: the offline setting is preferable when representations can be precomputed and the reader adapted, and the online setting when query-specific evidence selection, inspectable source text, and compatibility with unchanged downstream readers matter. Its online cost is one encoder pass per retrieved passage that jointly scores the passage and all its sentences.

\begin{figure}[h!]
    \centering
    \includegraphics[width=\linewidth]{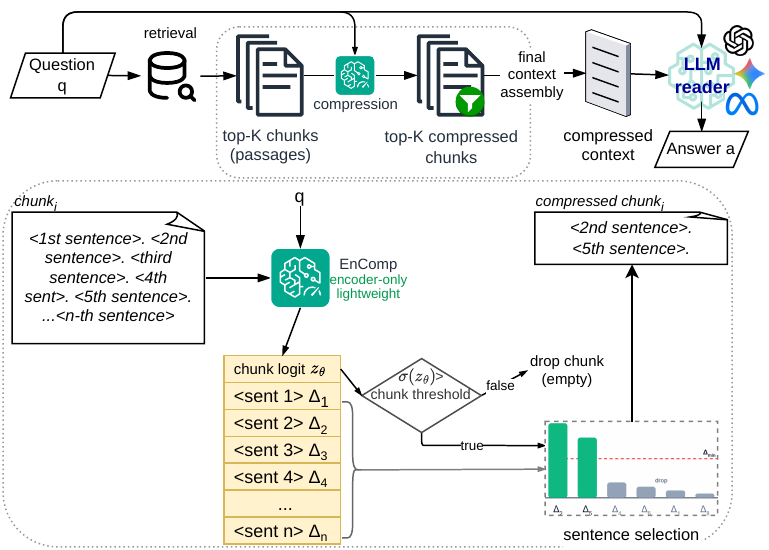}
    \caption{Overview of our framework. Our proposed lightweight context pruner includes three steps. (1) Model predicts the changes in clue richness, denoted as $\Delta$, assuming the sentence is omitted. A larger $\Delta$ indicates that the sentence is more critical. (2) Apply thresholdings at chunk level and sentence's deltas to retain most essential sentences. (3) Compressed chunks are joined in retrieved order to create final context.}
    \label{fig:main-pipeline}
\end{figure}

\section{Methodology}
Fig.~\ref{fig:main-pipeline} gives the overall inference flow of our compressor between the retriever and the reader. Given the user question and the retrieved chunks, EnComp compresses each chunk independently while conditioning on the question. For each chunk, a single encoder-only pass scores both the chunk and all of its sentences; these scores support the two-stage filtering shown in the figure, where chunks with little evidence are discarded and evidence-bearing sentences are retained from the remaining chunks. The retained sentences preserve their original order within each chunk, and the compressed chunks are then concatenated in retriever order to form the final context for the reader. We next formalize this compression setting, then describe amortized marginal contribution scoring, the training objective, and inference procedure.

\paragraph{Problem Formulation}
Given a query $q$ and top-$k$ retrieved passages $D_k = \langle d_1, \ldots, d_k \rangle$, a reader model $M$ generates an answer $y$ from the provided context. We aim to learn a compression function $\pi$ that produces a compact query-relevant context:
\[
    D'_k = \pi(q, D_k), \qquad \ell(D'_k)\ll \ell(D_k),
\]
where $\ell(\cdot)$ denotes context length, measured in tokens. The compressed context should preserve the evidence needed for $M$ to answer $q$.

\paragraph{RAG Pipeline with Compression}
Let $\mathcal{K}$ denote the external document corpus. For a query $q$, the retriever returns the top-$k$ passages $D_k=\langle d_1, \ldots, d_k \rangle=\text{Retriever}(q,\mathcal{K})$, which are then compressed as:
\[
    D'_k = \text{Compressor}(q, D_k), \quad \text{s.t. } \ell(D'_k) \ll \ell(D_k).
\]
The language model generates the final answer using $D'_k$ as context. Effective compression should retain evidence, remain lightweight, and substantially reduce token usage.

\subsection{Amortized Marginal Contribution Scoring}
For a passage $P=\langle s_1,\ldots,s_n \rangle$, spaCy \cite{honnibal2020spacy} provides the sentence boundaries, which are mapped to the encoded token sequence through a sentence-token mask indicating which tokens belong to each sentence. Contextual token representations within each sentence span are pooled into a sentence vector, giving the scoring head one aligned representation for each sentence.

For a query--passage pair $(q,P)$, our encoder-only pruner produces contextual representations for the full query--passage input. A pooling head summarizes the sequence into a global passage vector $g$, and a document head predicts an answerability logit:
\[
    z_\theta(q,P) = f_{\text{doc}}(g).
\]
For each sentence $s_i \in P$, the model obtains a contextual sentence vector $e_i$ from the same full-passage encoding and predicts a contribution score:
\[
    \hat{\Delta}_i = f_{\text{sent}}(e_i, g).
\]
The score $\hat{\Delta}_i$ is not an isolated relevance label; it estimates the marginal contribution of $s_i$, i.e., how much removing that sentence would change passage answerability. A high positive score therefore marks evidence whose removal would weaken the passage for answering $q$. This scoring is amortized: all sentence contributions are predicted from one full-context encoding, avoiding explicit per-sentence counterfactual re-encoding at inference time.


\subsection{Training Objective}
Each sentence has a binary label $y_i\in\{0,1\}$, where $1$ indicates a critical sentence. The passage-level target is
\[
    y_{\text{doc}} = \mathbb{I}\left[\exists i: y_i=1\right].
\]
The training signal follows the same marginal-contribution view: critical sentences should cause a larger answerability drop when removed than noncritical sentences. The model should therefore assign higher contribution scores to answer-critical evidence and separate them from distractors within the same passage.

\paragraph{Counterfactual Contribution Targets}
During training only, each sentence is paired with a counterfactual passage, i.e., the same passage with that sentence redacted. This gives a soft contribution target:
\[
    \Delta_i^\star =
    \operatorname{sg}\left(z_\theta(q,P) - z_\theta(q,P\setminus \{s_i\})\right),
\]
where $\operatorname{sg}(\cdot)$ denotes stop-gradient. This target captures the answerability drop caused by removing $s_i$. The sentence head is trained to approximate this counterfactual effect from the full-context encoding, so inference does not require evaluating counterfactual passages.

\paragraph{Marginal-Contrastive Objective}
Let $C=\{i:y_i=1\}$ and $N=\{j:y_j=0\}$. The main objective combines soft matching to marginal contribution targets with contrastive separation:
\[
\begin{aligned}
\mathcal{L}_{\text{mc}}
&= \lambda_{\text{dist}}\mathcal{L}_{\text{dist}}
 + \lambda_{\text{rank}}\mathcal{L}_{\text{rank}}, \\
\mathcal{L}_{\text{dist}}
&= \frac{1}{n}
   \sum_{i=1}^{n}
   \operatorname{SmoothL1}
   (\hat{\Delta}_i,\Delta_i^\star), \\
\mathcal{L}_{\text{rank}}
&= \frac{1}{|C||N|}
   \sum_{i\in C}\sum_{j\in N}
   \max\left(0,\mu-\hat{\Delta}_i+\hat{\Delta}_j\right).
\end{aligned}
\]
The $\mathcal{L}_{\text{dist}}$ calibrates each predicted contribution score against the observed answerability drop caused by removing that sentence, while the $\mathcal{L}_{\text{rank}}$ forces critical sentences to outrank noncritical ones in the same context. We set $\mathcal{L}_{\text{rank}}=0$ when either $C$ or $N$ is empty. This relative objective is more informative than independent binary classification because it directly models the competition between useful evidence and distracting context.

We further include a document-level BCE term $\mathcal{L}_{\text{doc}}$ on $z_\theta(q,P)$ to calibrate whether the passage contains any useful evidence. A lightweight zero-anchor regularizer $\mathcal{L}_{\text{zero}}$ (detailed in Appendix~\ref{subsec:training-config}) keeps noncritical sentence scores, and all sentence scores in clue-free passages, close to zero. The complete loss is:
\[
    \mathcal{L}
    = \mathcal{L}_{\text{mc}}
    + \lambda_{\text{doc}}\mathcal{L}_{\text{doc}}
    + \lambda_{\text{zero}}\mathcal{L}_{\text{zero}}.
\]

\subsection{Inference}
At inference time, the compressor segments each passage into sentences and runs the model once for the full query--passage pair. The document logit first acts as a coarse gate: if $\sigma(z_\theta(q,P))$ is below a tunable $d_{min}$ threshold, the passage is treated as non-informative. Otherwise, sentences are ranked by their predicted contribution scores $\hat{\Delta}_i$, and positive-scoring sentences above a tunable $\delta_{min}$ threshold are retained under a small retention budget. The selected sentences are returned in their original order to form the compressed context.

\section{Experiment}

\begin{table*}[!ht]
\centering
\setlength{\tabcolsep}{4.5pt}
\resizebox{0.985\linewidth}{!}{%
\begin{tabular}{l|cccc|cccc|cccc|cccc|cccc} 
\toprule[1.5pt]
              & \multicolumn{4}{c|}{NQ}                                       & \multicolumn{4}{c|}{TQA}                                      & \multicolumn{4}{c|}{HQA}                                      & \multicolumn{4}{c|}{2Wiki}                                    & \multicolumn{4}{c}{Musique}                                    \\ 
\cmidrule(l){2-21}
              & EM            & F1            & Lat           & Rate         & EM            & F1            & Lat           & Rate         & EM            & F1            & Lat           & Rate         & EM            & F1            & Lat           & Rate         & EM            & F1            & Lat           & Rate          \\ 
\midrule
\multicolumn{10}{l}{\textbf{Llama-3.1-8B Instruct}} & \multicolumn{11}{r}{\textit{Top-20 retrieved chunks}} \\ 
\midrule
\textit{Raw} & 37.2 & 49.3 & 0.000 & 100.0 & 64.0 & 72.7 & 0.000 & 100.0 & 31.1 & 41.7 & 0.000 & 100.0 & 25.6 & 32.0 & 0.000 & 100.0 & 4.6 & 11.1 & 0.000 & 100.0 \\
CompAct & 33.7 & 45.4 & 4.312 & \uline{3.7} & 60.0 & 69.0 & 4.102 & \uline{3.5} & 30.9 & 42.1 & 4.696 & \uline{3.7} & 21.6 & 28.2 & 4.905 & \uline{3.5} & 5.3 & 12.4 & 5.869 & \uline{4.1} \\
EXIT & 33.9 & 45.5 & 1.392 & 41.9 & 60.7 & 69.8 & 1.434 & 36.9 & 29.0 & 39.4 & 1.465 & 34.0 & 22.5 & 28.5 & 1.492 & 29.4 & 4.1 & 10.8 & 1.473 & 33.4 \\
RECOMP-abs & 32.9 & 44.9 & 1.223 & \textbf{2.1} & 55.9 & 65.3 & 1.332 & \textbf{2.0} & 28.5 & 39.8 & 1.206 & \textbf{1.8} & 22.4 & 28.2 & 1.145 & \textbf{1.4} & 5.2 & 11.9 & 1.393 & \textbf{1.6} \\
RECOMP-ext & 30.9 & 41.3 & \textbf{0.036} & 11.4 & 54.8 & 62.6 & \textbf{0.037} & 11.5 & 23.3 & 32.1 & \textbf{0.035} & 12.4 & 15.8 & 22.3 & \textbf{0.036} & 12.2 & 4.0 & 9.7 & \textbf{0.037} & 12.0 \\
Refiner & 31.0 & 42.0 & 6.727 & 8.8 & 59.1 & 68.0 & 5.013 & 6.5 & 29.1 & 39.3 & 3.915 & 4.4 & 22.2 & 28.1 & 3.342 & 3.7 & 5.1 & 11.6 & 4.551 & 5.0 \\
LongLLMLin & 31.7 & 43.8 & 1.098 & 30.1 & 62.6 & 71.2 & 1.108 & 30.0 & 30.8 & 41.3 & 1.104 & 30.0 & \uline{24.0} & \uline{29.9} & 1.133 & 30.2 & 4.9 & 11.3 & 1.120 & 30.3 \\
Provence & \uline{34.2} & \uline{46.2} & 0.188 & 20.8 & 62.3 & 71.1 & 0.189 & 18.9 & 30.0 & 40.5 & 0.199 & 13.8 & 22.5 & 28.6 & 0.197 & 11.9 & 5.3 & \uline{11.9} & 0.196 & 12.2 \\
AttnComp & 34.1 & 45.9 & 0.123 & 10.2 & \textbf{62.9} & \textbf{72.0} & 0.124 & 13.5 & \uline{32.0} & \uline{42.8} & 0.124 & 11.0 & 22.6 & 28.8 & 0.130 & 7.0 & \uline{5.5} & 11.8 & 0.126 & 4.8 \\
Ours & \textbf{36.2} & \textbf{48.6} & \uline{0.044} & 20.3 & \uline{62.6} & \uline{71.8} & \uline{0.045} & 18.5 & \textbf{33.2} & \textbf{44.4} & \uline{0.045} & 13.8 & \textbf{26.2} & \textbf{32.4} & \uline{0.046} & 11.5 & \textbf{5.9} & \textbf{13.7} & \uline{0.045} & 15.9 \\ 
\midrule
\multicolumn{10}{l}{\textbf{Llama-3.3-70B Instruct}} & \multicolumn{11}{r}{\textit{Top-20 retrieved chunks}} \\ 
\midrule
\textit{Raw} & 39.1 & 53.4 & 0.000 & 100.0 & 68.8 & 77.9 & 0.000 & 100.0 & 38.2 & 50.3 & 0.000 & 100.0 & 30.6 & 38.5 & 0.000 & 100.0 & 10.0 & 18.9 & 0.000 & 100.0 \\
CompAct & 35.2 & 48.5 & 4.312 & \uline{3.7} & 64.9 & 74.0 & 4.102 & \uline{3.5} & 36.1 & 47.8 & 4.696 & \uline{3.7} & 26.5 & 33.2 & 4.905 & \uline{3.5} & 8.4 & 15.9 & 5.869 & \uline{4.1} \\
EXIT & 36.0 & 49.9 & 1.392 & 41.9 & 65.8 & 75.3 & 1.434 & 36.9 & 35.0 & 46.6 & 1.465 & 34.0 & 25.7 & 32.9 & 1.492 & 29.4 & 8.0 & 17.0 & 1.473 & 33.4 \\
RECOMP-abs & 34.2 & 47.2 & 1.223 & \textbf{2.1} & 60.2 & 69.9 & 1.332 & \textbf{2.0} & 33.1 & 45.4 & 1.206 & \textbf{1.8} & 27.7 & 33.4 & 1.145 & \textbf{1.4} & 6.7 & 14.8 & 1.393 & \textbf{1.6} \\
RECOMP-ext & 33.5 & 45.7 & \textbf{0.036} & 11.4 & 64.0 & 72.6 & \textbf{0.037} & 11.5 & 28.3 & 38.9 & \textbf{0.035} & 12.4 & 17.1 & 24.6 & \textbf{0.036} & 12.2 & 6.0 & 14.1 & \textbf{0.037} & 12.0 \\
Refiner & 33.4 & 46.4 & 6.727 & 8.8 & 64.1 & 73.6 & 5.013 & 6.5 & 33.4 & 45.0 & 3.915 & 4.4 & 25.7 & 32.0 & 3.342 & 3.7 & 7.8 & 16.1 & 4.551 & 5.0 \\
LongLLMLin & 34.5 & 48.8 & 1.098 & 30.1 & \uline{67.3} & 76.6 & 1.108 & 30.0 & 36.3 & 48.7 & 1.104 & 30.0 & 28.7 & 36.2 & 1.133 & 30.2 & 8.5 & 17.5 & 1.120 & 30.3 \\
Provence & 37.0 & 51.3 & 0.188 & 20.8 & 66.6 & 76.0 & 0.189 & 18.9 & 35.4 & 47.5 & 0.199 & 13.8 & 26.1 & 32.8 & 0.197 & 11.9 & 8.5 & 17.8 & 0.196 & 12.2 \\
AttnComp & \uline{37.6} & \uline{51.3} & 0.123 & 10.2 & \textbf{68.1} & \textbf{77.7} & 0.124 & 13.5 & \uline{37.9} & \uline{50.6} & 0.124 & 11.0 & \textbf{32.7} & \textbf{39.5} & 0.130 & 7.0 & \uline{9.6} & \uline{19.0} & 0.126 & 4.8 \\
Ours & \textbf{37.7} & \textbf{52.2} & \uline{0.044} & 20.3 & 67.0 & \uline{76.8} & \uline{0.045} & 18.5 & \textbf{38.1} & \textbf{51.0} & \uline{0.045} & 13.8 & \uline{32.2} & \uline{38.4} & \uline{0.046} & 11.5 & \textbf{9.9} & \textbf{19.2} & \uline{0.045} & 15.9 \\
\bottomrule[1.5pt]
\end{tabular}
}
\caption{Performance across models and datasets by readers Llama-3.1-8B Instruct and Llama-3.3-70B Instruct with $k=20$. Best (\textbf{bold}) and second-best (\uline{underlined}) are determined using higher-precision values before rounding. Compression Latency (Lat$\downarrow$) is in seconds, and Compression Ratio (Rate$\downarrow$) is reported as a percentage.}
\label{tab:main}
\end{table*}

\subsection{Experimental Setup}

\paragraph{Training Phase}
Similar to existing works such as \cite{hwang2024exit, luo-etal-2025-attncomp}, we use a random subset 30\% out of the \textit{original} HotpotQA \cite{yang2018hotpotqa} train set to train with a 9:1 split, where there are passages (and annotated critical sentences) for each question. We merged very short passages into longer ones to mitigate the dataset's severe passage-length imbalance, shorten training time, and enable additional data augmentation. Training is performed on a single RTX 4090 GPU using LoRA \cite{hu2022lora} to finetune the backbone with the standard separator token (e.g., [SEP]) automatically defined.

\paragraph{RAG Configuration}
The main pipeline consists of 3 primary components. The Contriever-MSMARCO \cite{izacard2021unsupervised} was used as chunk retriever on the \textit{2018 Wikipedia corpus} dataset \cite{karpukhin2020dense} to create passages for each question from Question Answering (QA) datasets. Our compressor is based on an encoder-only model -- ModernBERT \cite{warner2024smarter}, large version (395M) used as default, base (139M) used in ablation. 

For reading to answer, we mainly employed Llama-3.1-8B-Instruct, Llama-3.3-70B-Instruct \cite{dubey2024llama}; and with proprietary Google Gemini-3-flash \cite{google2025gemini3flash}, OpenAI GPT-5-mini \cite{openai_gpt5mini_2025} for further exploration. Vertex AI, OpenRouter, OpenAI API services and self-hosted vLLM \cite{kwon2023efficient} were used to provide the mentioned reader APIs.

To evaluate, we use HotpotQA (HQA) \cite{yang2018hotpotqa}, 2WikiMultihopQA (2WIKI) \cite{ho2020constructing}, and Musique \cite{trivedi2022musique} datasets as multi-hop QA benchmarks; and Natural Questions (NQ) \cite{kwiatkowski2019natural}, TriviaQA (TQA) \cite{joshi2017triviaqa} datasets as single-hop QA benchmarks. The evaluated sets are the same as \cite{yoon2024compact,hwang2024exit}.

\paragraph{Baselines}
We compare our method with eight context compression baselines: 
\textbf{RECOMP-Abs}/\textbf{RECOMP-Ext} \cite{xu2024recomp}, abstractive and extractive variants; 
\textbf{CompAct} \cite{yoon2024compact}, an iterative Mistral-7B compressor; 
\textbf{Refiner} \cite{li2024refiner}, based on Llama2-7B; 
\textbf{LongLLMLingua} \cite{jiang2023longllmlingua} (LongLLMLin), based on Llama2-7B; 
\textbf{EXIT} \cite{hwang2024exit}, based on Gemma-2B; 
\textbf{Provence} \cite{chirkova2025provence}, based on DeBERTaV3; 
and \textbf{AttnComp} \cite{luo-etal-2025-attncomp}, based on Llama-3.1-8B Instruct. 
For a fair comparison, all baselines use the same preprocessing pipeline, FlashAttention-2 when available \cite{dao2023flashattention2}, and the same thread count. We also include an uncompressed \textit{Raw} baseline in the main results. 

\paragraph{Evaluation Metrics}
Exact Match (EM$\uparrow$) and F1$\uparrow$ score are used to measure accuracy in question answering. Compression latency (Lat$\downarrow$, in seconds), compression ratio (Rate$\downarrow$) are used to evaluate compression speed and compactness. The latency here is the compressed time measured in an end-to-end manner to simulate a realistic processing timeline. The compression rate is defined as the ratio of the length of the compressed sequence to the length of the original sequence using $cl100k\_base$ tokenizer. Each experiment is run 3 times (different batch sizes) on a single RTX 4090 GPU and the average is reported.

\subsection{Main Results}

Tab.~\ref{tab:main} shows evaluation results across five datasets using two commonly used readers at $k=20$. Note that the compression latency and rate depend on the top-$k$ value and are the same regardless of the reader. Although trained only on HQA questions with the original HQA context corpus, our pruner generalizes well across datasets, covering both single-hop and multi-hop settings. Compared with other compressors, our method mostly yields the best or second-best EM/F1 metrics across datasets and readers. It also achieves approximately equal answer quality to, or surpasses, the \textit{Raw} baseline in several cases. Regarding compression efficiency, our method remains highly responsive--achieving the second-best latency among compressors with less than $0.05$s and retains only $11$--$20\%$ of the original context.

Meanwhile, existing methods often struggle to maintain a good trade-off across answer quality, compression rate, and latency. For example, RECOMP-ext is the fastest compressor but usually yields much lower answer quality. Similarly, CompAct or RECOMP-abs compresses context to a very compact level, but at the cost of slow execution. A strong baseline such as AttnComp achieves comparable EM/F1 but is still $\sim 2.75\times$ slower than ours.

\begin{table}[h]
        \centering
        \resizebox{0.85\linewidth}{!}{%
        \begin{tabular}{lrrrr} 
        \toprule[1.5pt]
\textbf{Method} & \textbf{EM} $\uparrow$ & \textbf{F1} $\uparrow$ & \textbf{Lat} $\downarrow$ & \textbf{Rate} $\downarrow$ \\
        \midrule
CompAct & 32.2 & 41.5 & 3.670 & \uline{7.7} \\
EXIT & 31.2 & 40.5 & 0.896 & 37.2 \\
RECOMP-abs & 30.4 & 39.7 & 0.910 & \textbf{4.3} \\
RECOMP-ext & 29.1 & 38.0 & \textbf{0.023} & 30.4 \\
Refiner & 31.6 & 40.9 & 3.230 & 8.5 \\
LongLLMLin & 31.5 & 40.8 & 0.682 & 33.5 \\
Provence & 32.3 & 41.9 & 0.126 & 17.9 \\
AttnComp & \uline{33.8} & \uline{43.4} & 0.080 & 16.5 \\
Ours & \textbf{34.3} & \textbf{44.1} & \uline{0.029} & 19.2 \\
        \bottomrule[1.5pt]
        \end{tabular}
        }
    \caption{Overall performance across compressors, via 2-stage averaging: first over 4 settings of 2 readers $\times$ 2 retrieved depths ($k=5,20$), then over all five datasets.}
    \label{tab:overall}
\end{table}

For an overall view, we aggregate main experimental results into a single metric vector per compressor as in Tab.~\ref{tab:overall}. For each metric, we take unweighted means in 2 steps: (i) average across the 4 experimental settings -- 2 readers $\times$ 2 retrieval depths $k=5,20$; then (ii) average across the 5 datasets. In general, we achieve among the best results on answer accuracy while holding the second-fastest position and a decent compactness ratio. Results at $k=5$ retrieval are shown in Tab.~\ref{tab:k5} in the Appendix.

In Tab. \ref{tab:gemini} in the Appendix, we conducted experiments with proprietary Gemini-3-flash and GPT-5-mini at a low thinking budget as readers on HQA (intra-domain) and 2Wiki (inter-domain) at retrieval depth $k=20$ for further comparison. Results show a similar pattern and confirm the efficiency: we mostly achieve the best or second-best answer quality with the fastest latency in the table.

\begin{figure*}[t]
    \centering
        \includegraphics[width=\linewidth]{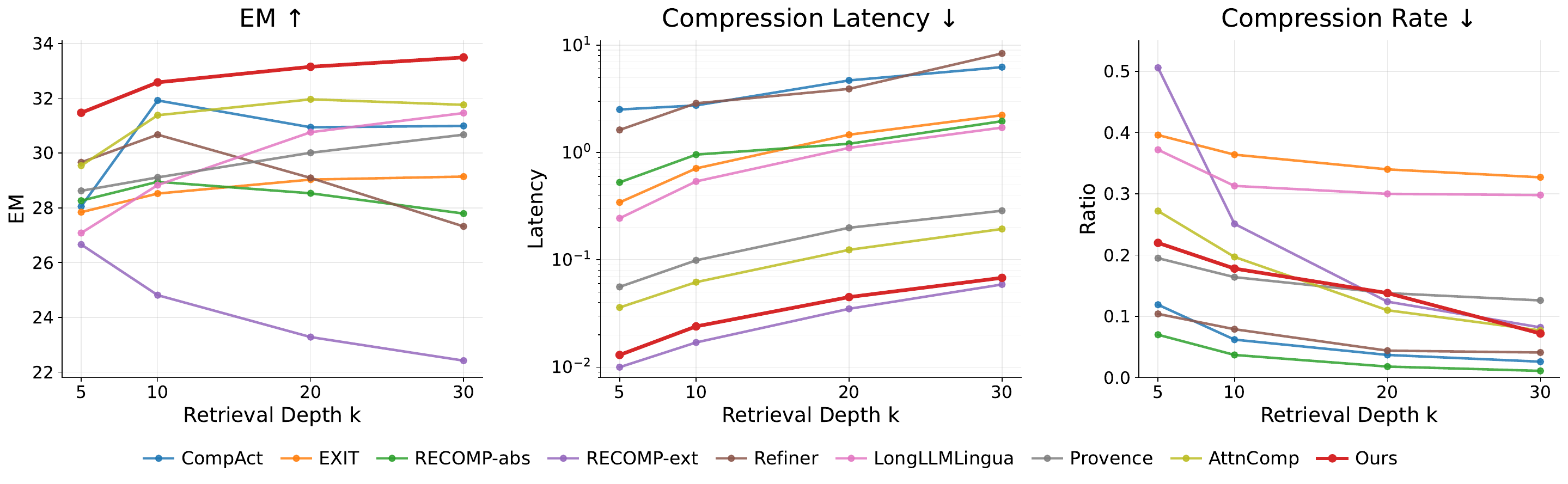}
        \caption{Analysis on HQA at increasing top-k $=\{5, 10, 20, 30\}$ by reader Llama-3.1-8B Instruct, comparing EM, and compression latency between baselines and our proposed method.}
        \label{fig:line-chart}
\end{figure*}

\paragraph{Robustness}
To assess robustness as the retrieved set enlarges, we vary the retrieval depth $k \in \{5, 10, 20, 30\}$ and report the results in Fig.~\ref{fig:line-chart}. Our method shows stable gains as $k$ increases, with EM improving from 31.47 at $k=5$ to 33.49 at $k=30$. It consistently achieves competitive EM across all retrieval depths, while some baselines exhibit unstable or degraded performance when more retrieved chunks are introduced. In terms of efficiency, our method maintains low compression latency, increasing moderately from 0.013s to 0.068s as $k$ grows. Although RECOMP-ext is slightly faster, its EM drops substantially at larger $k$, whereas our method preserves both high accuracy and low latency. The compression-rate results further show that our method keeps a compact representation without relying on overly aggressive pruning. Overall, these results suggest that our approach scales robustly with retrieval depth while maintaining a favorable balance among accuracy, latency, and compression compactness.

To visualize model comparisons at $k=20$, Fig.~\ref{fig:tradeoff} summarizes the quality--latency--memory trade-off across five datasets using the Llama-3.3-70B-Instruct reader. Our method achieves comparable EM to the best while requiring only 1.86 GB peak VRAM and 0.045s compression latency, making it substantially more efficient than the strongest accuracy baseline, which requires 3.7$\times$ more VRAM and nearly 3$\times$ higher latency. This efficiency is particularly important for resource-constrained deployment, where edge-class devices often operate under tight few-GB shared-memory budgets. These results show that our method preserves high answering accuracy while remaining practical for memory- and latency-sensitive scenarios.

\section{Ablation Study}

To assess the training objective, we compare the default model with a variant trained using direct binary classification at each sentence head. As shown in Tab.~\ref{tab:model-ablation} with both readers (Llama-3.1-8B Instruct and Llama-3.3-70B Instruct), the default model achieves the better or mostly tied performance on most metrics across the five datasets. The binary-loss variant (the 2nd row) remains competitive, with identical latency and a slightly more compact output, but generally performs slightly worse, except on HQA where the results are similar. This suggests that the proposed loss $\mathcal{L}$ provides a consistently better generalization.

We also study backbone capacity by replacing ModernBERT-large with ModernBERT-base one while keeping $\mathcal{L}$ unchanged. The base variant (the 3rd row in Tab.~\ref{tab:model-ablation}) performs lower across all reported metrics, indicating the benefit of the larger backbone. However, it is approximately 30\% faster (around RECOMP-ext speed) in our measurements, although with a slightly less compact output, showing a clear trade-off between quality and speed.

\begin{table}[h]
\centering
\setlength{\tabcolsep}{3pt}
\resizebox{\linewidth}{!}{%
\begin{tabular}{l|cc|cc|cc|cc|cc}
\toprule[1.5pt]
\multirow{2}{*}{Variants} 
& \multicolumn{2}{c|}{NQ} 
& \multicolumn{2}{c|}{TQA} 
& \multicolumn{2}{c|}{HQA} 
& \multicolumn{2}{c|}{2Wiki} 
& \multicolumn{2}{c}{Musique} \\
\cmidrule(r){2-11}
& EM & F1 
& EM & F1 
& EM & F1 
& EM & F1 
& EM & F1 \\
\midrule
\multicolumn{11}{r}{\textbf{Llama-3.1-8B Instruct}} \\
\midrule
\textit{default}            & 36.2 & 48.6 & 62.6 & 71.8 & 33.2 & 44.4 & 26.2 & 32.4 & 5.9 & 13.7 \\
\specialrule{0.25pt}{1pt}{1pt}
-large \& $\mathcal{L}_{bin}$ & 35.3 & 47.9 & 62.2 & 70.6 & 33.2 & 44.5 & 25.6 & 31.7 & 5.3 & 12.3 \\
\specialrule{0.25pt}{1pt}{1pt}
-base \& $\mathcal{L}$      & 33.8 & 46.3 & 61.3 & 70.7 & 31.9 & 43.0 & 25.5 & 31.5 & 5.1 & 12.4 \\
\midrule
\multicolumn{11}{r}{\textbf{Llama-3.3-70B Instruct}} \\
\midrule
\textit{default}            & 37.7 & 52.2 & 67.0 & 76.8 & 38.1 & 51.0 & 32.2 & 38.4 & 9.9 & 19.2 \\
\specialrule{0.25pt}{1pt}{1pt}
-large \& $\mathcal{L}_{bin}$ & 36.9 & 51.6 & 66.8 & 76.1 & 38.2 & 51.2 & 31.7 & 38.3 & 9.5 & 19.0 \\
\specialrule{0.25pt}{1pt}{1pt}
-base \& $\mathcal{L}$      & 34.8 & 49.1 & 66.5 & 75.9 & 37.7 & 49.8 & 31.0 & 37.0 & 8.7 & 18.9 \\
\bottomrule[1.5pt]
\end{tabular}
}
\caption{Performance across 5 datasets on our model trained with the direct binary classification (at each sentence head), and a small variant using ModernBERT-base backbone (still with proposed loss $\mathcal{L}$) at $k=20$, evaluated with two readers. \textit{default} uses -large backbone with the loss $\mathcal{L}$.}
\label{tab:model-ablation}
\end{table}

\begin{table}[h]
        \centering
        \resizebox{0.9\linewidth}{!}{%
        \begin{tabular}{lrrrr} 
        \toprule[1.5pt]
\textbf{Method} & \textbf{EM} $\uparrow$ & \textbf{F1} $\uparrow$ & \textbf{Lat} $\downarrow$ & \textbf{Rate} $\downarrow$ \\
        \midrule
        \multicolumn{5}{r}{\textbf{BM25}} \\
        \midrule
CompAct & 31.8 & 42.9 & 2.787 & \uline{6.5} \\
EXIT & 29.8 & 40.8 & 0.697 & 36.6 \\
RECOMP-abs & 29.9 & 42.4 & 0.698 & \textbf{3.6} \\
RECOMP-ext & 28.2 & 37.3 & \textbf{0.016} & 27.3 \\
Refiner & 32.4 & 43.3 & 2.750 & 8.3 \\
LongLLMLin & 30.4 & 40.7 & 0.520 & 31.3 \\
Provence & 30.0 & 41.0 & 0.099 & 19.0 \\
AttnComp & \uline{33.6} & \uline{44.7} & 0.061 & 21.1 \\
Ours & \textbf{34.2} & \textbf{46.2} & \uline{0.023} & 15.4 \\
        \midrule
        \multicolumn{5}{r}{\textbf{BGE-M3}} \\
        \midrule
EXIT & 31.6 & 41.8 & 0.692 & 38.6 \\
RECOMP-abs & 31.8 & 44.4 & 0.695 & \textbf{3.9} \\
RECOMP-ext & 29.0 & 39.8 & \textbf{0.017} & 27.0 \\
LongLLMLin & 31.9 & 42.8 & 0.529 & 31.4 \\
Provence & 34.5 & 46.5 & 0.090 & \uline{19.5} \\
AttnComp & \uline{35.2} & \uline{46.8} & 0.061 & 24.6 \\
Ours & \textbf{36.1} & \textbf{48.4} & \uline{0.023} & 20.0 \\
        \bottomrule[1.5pt]
        \end{tabular}
        }
    \caption{Performance across compressors using Llama-3.1-8B Instruct as the reader at $k=10$ with BM25 and BGE-M3 retrievers on a subset of 500 HotpotQA examples.}
    \label{tab:bm25}
\end{table}

\paragraph{With Different Retrievers}
We further examine whether the behavior of different compressors remains consistent when the upstream retriever changes. Specifically, we evaluate the compressors on HotpotQA using both a sparse retriever, BM25 \cite{robertson2009probabilistic}, and a dense multilingual retriever, BGE-M3 \cite{chen-etal-2024-m3}. As shown in Tab.~\ref{tab:bm25}, our method performs competitively under both retrieval settings and achieves marginally higher EM and F1 than the strongest baselines in this experiment. The gains over AttnComp are relatively small, but they are observed with both retrievers, suggesting that the proposed compression strategy can preserve useful evidence across different retrieval distributions.

The results also highlight trade-offs among answer accuracy, compression rate, and inference time. RECOMP-ext is the fastest method, but its downstream accuracy is substantially lower than that of stronger compressors. RECOMP-abs produces the lowest compression rate, yet this does not lead to better accuracy or latency. Among the faster baselines, AttnComp obtains relatively high EM and F1, providing a useful comparison for evaluating efficiency-oriented compression. Our method shows a modest improvement in EM and F1 while keeping inference time low. Overall, these results suggest that the proposed method offers a balanced effectiveness-efficiency trade-off across both BM25 and BGE-M3 retrieval settings.

\paragraph{With Different Corpus}
The main experiments above are conducted on standard settings built over the \textit{2018 Wikipedia corpus} \cite{karpukhin2020dense}. To further examine the behavior of different compressors beyond this source distribution, we additionally evaluate them on QASPER \cite{dasigi-etal-2021-dataset}, where the evidence corpus consists of NLP academic papers rather than Wikipedia passages, while keeping the same retriever, chunk size, and $k=20$ setting. As shown in Table~\ref{tab:qasper-compressors}, our method remains highly competitive on this specialized corpus, achieving the second-best EM and F1 while also obtaining the second-lowest latency. This indicates that the method can preserve useful evidence for answer prediction without incurring the high processing cost observed in several baselines. Although some methods optimize more strongly for either compression rate or latency alone, their answer quality is less consistent in this setting. Overall, the results suggest that our method offers a favorable and practical trade-off across accuracy, latency, and compression compactness under a non-Wikipedia academic corpus.

\begin{table}[h]
\centering
\resizebox{0.9\linewidth}{!}{%
\begin{tabular}{lcccc}
\toprule[1.5pt]
\textbf{Method} & \textbf{EM} $\uparrow$ & \textbf{F1} $\uparrow$ & \textbf{Lat} $\downarrow$ & \textbf{Rate} $\downarrow$ \\
\midrule
EXIT          & 9.1  & 21.1 & 1.326 & 38.5 \\
RECOMP-abs    & 8.7  & 16.7 & 1.222 & \textbf{1.9} \\
RECOMP-ext    & 9.2  & 16.5 & \textbf{0.035} & \uline{13.2} \\
LongLLMLing   & 8.5  & 19.8 & 1.053 & 31.6 \\
Provence      & \textbf{10.0} & \textbf{22.5} & 0.157 & 24.3 \\
AttnComp      & 9.8  & 21.1 & 0.123 & 16.2 \\
Ours          & \uline{9.9}  & \uline{21.3} & \uline{0.043} & 24.2 \\
\bottomrule[1.5pt]
\end{tabular}
}
\caption{Performance across compressors on QASPER (dev set), a non-Wikipedia NLP academic corpus using the same chunk size and retriever, at $k=20$ using the Llama-3.3-70B reader.}
\label{tab:qasper-compressors}
\end{table}

\section{Conclusion}
We introduced \textbf{EnComp}, a lightweight encoder-only framework for query-driven context compression in RAG. EnComp scores sentences from a single full-context encoding and trains the scorer with marginal contribution signals and contrastive ranking, keeping compression extractive while avoiding generative rewriting. Across single-hop and multi-hop QA benchmarks, multiple readers, retrieval depths, and retriever settings, EnComp achieves strong answer quality while maintaining low compression latency and under 2 GB peak compressor memory. Rather than aiming to dominate every aspect, EnComp offers a practical trade-off between answer fidelity, compactness, efficiency, and implementation simplicity. These results suggest that encoder-only sentence pruning can serve as a useful alternative to larger compression pipelines, especially for resource-constrained RAG scenarios where stable performance and low overhead are both important.

\section{Limitations}
Our approach relies on sentence-level annotations for training. In this work, such annotations are available from the HQA dataset, but preparing similar labels for new domains may still require additional effort. Although strong large language models could help generate such labels, this may introduce extra cost and noise.
Another limitation is that EnComp compresses context at the sentence level. This keeps the method simple and efficient, but some unnecessary words may remain when useful and irrelevant information appear in the same sentence. Finer-grained pruning at the phrase or clause level may improve compactness, but could also increase complexity. Moreover, because EnComp selects evidence specifically for each query, compression is performed at query time rather than precomputed offline, an option available to query-independent soft compression when the corpus is fixed and the reader can be adapted. Although our evaluation spans single-hop, multi-hop, and academic-paper QA, it does not yet establish performance on long-form answer generation. Finally, the document-level and sentence-level thresholds may need light calibration for different readers, retrievers, or application budgets.

\section*{Acknowledgments}
This work was supported by the IITP (Institute of Information \& Coummunications Technology Planning \& Evaluation) - ITRC (Information Technology Research Center) grant funded by the Korea government(Ministry of Science and ICT)(IITP-2026-RS-2023-00259703).

This work was also supported by the National Research Foundation of Korea (NRF) grant funded by the Korea government (MSIT)(RS-2025-00573160).
This research was supported by the "Advanced GPU Utilization Support Program" funded by the Government of the Republic of Korea (Ministry of Science and ICT).

The first author was also supported by Hyundai Motor Chung Mong-Koo Global Scholarship and we are also thankful for the research credits from Google Cloud Platform.

\clearpage
\bibliography{custom}

\clearpage
\appendix

\section*{Appendix}
In the Appendix, additional implementation details, supplementary results and analyses which are not covered in the main text is provided.

\section{Extended Implementation Details}
This section reports our training environment and prompt templates. All training experiments are done on single NVIDIA RTX 4090 GPUs (multiple similar workstations: same GPU, similar CPUs and RAM sizes).

\subsection{Training Configuration}
\label{subsec:training-config}
We adopt LoRA \cite{hu2022lora} to finetune the pruning model at bfloat16 precision for parameter-efficient training while preserving generalization from the original weights. The model was trained with the following hyper-parameters:
\begin{itemize}[itemsep=0.01em]
    \item Batch size: $\{16, 32\}$
    \item Gradient accumulation steps: 8
    \item Learning rate: 7e-5
    \item Weight decay: 0.02
    \item Warmup steps: 200
    \item Training epochs: up to 6
    \item Optimizer: AdamW
    \item LoRA configuration: Rank = 64, Scaling = 16, Dropout = 0.1
    \item Loss hyperparameters: $\mu=0.35$, $\lambda_{\text{dist}}=1.0$, $\lambda_{\text{rank}}=1.5$, $\lambda_{\text{zero}}=1.0$, $\lambda_{\text{doc}}=0.6$; document BCE positive-class weight: 5.0
\end{itemize}

\paragraph{Zero-anchor regularizer}
$\mathcal{L}_{\text{zero}}$ is a minor calibration regularizer. It applies $\operatorname{SmoothL1}(\hat{\Delta}_i,0)$ to noncritical sentences. This discourages clearly unhelpful sentences from receiving positive contribution scores.

Regarding the detailed model architecture, our model is a lightweight scoring architecture built on a pretrained ModernBERT encoder plus a learnable multi-head attention pooling layer. Instead of CLS or mean pooling, H learned query vectors (one per pooling head; H = number of pooling heads $=8$, with hidden size divisible by H) attend over the token sequence to produce head-specific summaries that are concatenated and linearly projected. Padding is masked before and after softmax for stability. A dropout layer and a final linear unit map the pooled representation to a single scalar score.

Model selection was guided by validation loss and it took around 7 hours for each training (up to 6 epochs).

\subsection{Data Augmentation \& Pre-processing.}
We also add some data augmentation operations: \textbf{(1)} Randomly drop extra noncritical sentences when inspecting (dropping) a critical sentence: 10\%; \textbf{(2)} Randomly drop extra noncritical sentences when inspecting (dropping) a noncritical sentence: 10\%; \textbf{(3)} Randomly insert punctuation between sentences: 20\%; \textbf{(4)} Randomly add start word(phrase), end word(phrase): 5\%. We only used the query and the corresponding context set as the \textbf{original} HotpotQA set provided; no cross-pairing (i.e., a query with a non-corresponding context set) was used.

Some pre-processing we used: \textbf{(1)} merge overlapped chunks from a same document to reduce duplication in parts and titles (not longer than 20 sentences); \textbf{(2)} too short chunks--having less than 4 sentences, were also merged to become longer chunks for less extreme length imbalance in dataset (not longer than 12 sentences).

\subsection{More Inference Configuration}
Document-level threshold and delta min $d_{min}, \delta_{\min}$ were determined by grid search for each model weight set as mentioned before. Our best result checkpoint is configured with around $d_{min}=0.12, \delta_{\min}=0.01$

\subsection{QA Prompt Template}
Below is the QA prompt template for LLM readers to consider the query with the provided compressed context and answer.

\begin{lstlisting}[breaklines=true, basicstyle=\ttfamily\footnotesize, frame=tb, label=lst:prompt-template]
Context information is:
```{Merged Compressed Context}```

Given provided context (might not be sufficient for below query), answer the query without any explanation.
Query: `{question}`
Answer (in plain text):
\end{lstlisting}

\section{Additional Experimental Results}

\begin{table*}[!ht]
\centering
\setlength{\tabcolsep}{4.5pt}
\resizebox{0.985\linewidth}{!}{%
\begin{tabular}{l|cccc|cccc|cccc|cccc|cccc} 
\toprule[1.5pt]
              & \multicolumn{4}{c|}{NQ}                                       & \multicolumn{4}{c|}{TQA}                                      & \multicolumn{4}{c|}{HQA}                                      & \multicolumn{4}{c|}{2Wiki}                                    & \multicolumn{4}{c}{Musique}                                    \\ 
\cmidrule(l){2-21}
              & EM            & F1            & Lat           & Rate         & EM            & F1            & Lat           & Rate         & EM            & F1            & Lat           & Rate         & EM            & F1            & Lat           & Rate         & EM            & F1            & Lat           & Rate          \\ 
\midrule
\multicolumn{10}{l}{\textbf{Llama-3.1-8B Instruct}} & \multicolumn{11}{r}{\textit{Top-5 retrieved chunks}} \\ 
\midrule
\textit{Raw} & 35.4 & 47.3 & 0.000 & 100.0 & 60.9 & 69.8 & 0.000 & 100.0 & 30.2 & 40.9 & 0.000 & 100.0 & 23.5 & 30.2 & 0.000 & 100.0 & 4.3 & 10.8 & 0.000 & 100.0 \\
CompAct & 31.4 & 42.0 & 2.573 & 12.1 & 60.0 & 69.1 & 2.594 & 12.0 & 28.1 & 37.5 & 2.517 & 11.9 & 22.1 & 28.8 & 2.354 & 10.6 & \textbf{6.1} & \textbf{13.6} & 2.781 & 11.9 \\
EXIT & 31.3 & 42.6 & 0.332 & 45.8 & 58.1 & 66.9 & 0.335 & 41.6 & 27.8 & 38.2 & 0.343 & 39.6 & 20.7 & 26.5 & 0.350 & 33.4 & 4.1 & 10.3 & 0.346 & 36.2 \\
RECOMP-abs & 33.5 & 45.1 & 0.765 & \textbf{7.8} & 53.8 & 63.5 & 0.578 & \textbf{7.6} & 28.3 & 39.4 & 0.527 & \textbf{7.0} & 24.4 & 29.6 & 0.436 & \textbf{5.5} & 4.3 & 11.8 & 0.499 & \textbf{6.6} \\
RECOMP-ext & 33.8 & 45.1 & \textbf{0.009} & 47.4 & 57.4 & 65.7 & \textbf{0.009} & 47.7 & 26.7 & 36.5 & \textbf{0.010} & 50.6 & 19.2 & 25.9 & \textbf{0.008} & 50.1 & 4.2 & 10.5 & \textbf{0.009} & 48.6 \\
Refiner & 33.2 & 44.9 & 2.403 & 15.3 & 60.3 & 69.4 & 1.967 & 13.2 & 29.7 & 40.5 & 1.624 & 10.4 & 22.2 & 28.2 & 1.325 & 8.3 & 5.3 & 12.0 & 1.436 & 8.9 \\
LongLLMLin & 28.3 & 39.4 & 0.258 & 37.2 & 58.6 & 67.0 & 0.253 & 36.5 & 27.1 & 37.1 & 0.244 & 37.2 & 21.3 & 26.8 & 0.253 & 37.3 & 4.0 & 10.3 & 0.245 & 36.4 \\
Provence & 33.4 & 44.8 & 0.059 & 26.4 & 59.9 & 69.1 & 0.056 & 25.4 & 28.6 & 39.5 & 0.056 & 19.5 & 21.4 & 27.6 & 0.059 & 15.5 & 4.5 & 11.2 & 0.056 & 14.7 \\
AttnComp & 33.3 & 45.0 & 0.036 & 27.1 & \textbf{61.2} & \textbf{70.2} & 0.035 & 30.3 & 29.5 & 40.0 & 0.036 & 27.2 & 21.7 & 28.0 & 0.036 & 21.1 & 4.7 & 10.9 & 0.036 & 12.9 \\
Ours & \textbf{34.4} & \textbf{46.5} & 0.012 & 27.7 & 60.8 & 69.9 & 0.013 & 25.4 & \textbf{31.5} & \textbf{42.3} & 0.013 & 22.0 & \textbf{24.9} & \textbf{30.8} & 0.013 & 17.5 & 5.5 & 12.8 & 0.013 & 19.8 \\ 
\midrule
\multicolumn{10}{l}{\textbf{Llama-3.3-70B Instruct}} & \multicolumn{11}{r}{\textit{Top-5 retrieved chunks}} \\ 
\midrule
\textit{Raw} & 36.8 & 51.1 & 0.000 & 100.0 & 65.8 & 75.2 & 0.000 & 100.0 & 35.6 & 47.8 & 0.000 & 100.0 & 27.2 & 34.9 & 0.000 & 100.0 & 8.2 & 16.8 & 0.000 & 100.0 \\
CompAct & 35.3 & 48.4 & 2.573 & 12.1 & 65.2 & 74.6 & 2.594 & 12.0 & 36.6 & 48.6 & 2.517 & 11.9 & 27.0 & 34.6 & 2.354 & 10.6 & 8.8 & 17.4 & 2.781 & 11.9 \\
EXIT & 33.4 & 46.3 & 0.332 & 45.8 & 64.1 & 73.5 & 0.335 & 41.6 & 33.4 & 44.9 & 0.343 & 39.6 & 23.5 & 30.0 & 0.350 & 33.4 & 7.1 & 15.6 & 0.346 & 36.2 \\
RECOMP-abs & 33.9 & 46.6 & 0.765 & \textbf{7.8} & 56.8 & 66.8 & 0.578 & \textbf{7.6} & 32.2 & 44.3 & 0.527 & \textbf{7.0} & 26.7 & 32.3 & 0.436 & \textbf{5.5} & 6.5 & 14.2 & 0.499 & \textbf{6.6} \\
RECOMP-ext & 36.4 & 49.1 & \textbf{0.009} & 47.4 & 65.0 & 74.1 & \textbf{0.009} & 47.7 & 32.7 & 44.1 & \textbf{0.010} & 50.6 & 22.0 & 30.0 & \textbf{0.008} & 50.1 & 7.0 & 15.3 & \textbf{0.009} & 48.6 \\
Refiner & 35.1 & 48.6 & 2.403 & 15.3 & 65.6 & 75.2 & 1.967 & 13.2 & 35.8 & 47.6 & 1.624 & 10.4 & 25.7 & 32.7 & 1.325 & 8.3 & 8.3 & 16.6 & 1.436 & 8.9 \\
LongLLMLin & 31.7 & 44.6 & 0.258 & 37.2 & 65.2 & 74.2 & 0.253 & 36.5 & 32.5 & 44.2 & 0.244 & 37.2 & 24.8 & 31.4 & 0.253 & 37.3 & 7.6 & 16.2 & 0.245 & 36.4 \\
Provence & 35.7 & 49.8 & 0.059 & 26.4 & 66.1 & 75.6 & 0.056 & 25.4 & 35.1 & 47.4 & 0.056 & 19.5 & 27.0 & 33.7 & 0.059 & 15.5 & 6.8 & 15.9 & 0.056 & 14.7 \\
AttnComp & \textbf{36.5} & \textbf{50.2} & 0.036 & 27.1 & \textbf{67.7} & \textbf{77.1} & 0.035 & 30.3 & 36.8 & 49.3 & 0.036 & 27.2 & \textbf{32.4} & \textbf{39.3} & 0.036 & 21.1 & 8.6 & 18.2 & 0.036 & 12.9 \\
Ours & 36.2 & 50.1 & 0.012 & 27.7 & 66.4 & 75.8 & 0.013 & 25.4 & \textbf{37.0} & \textbf{49.6} & 0.013 & 22.0 & 32.1 & 38.1 & 0.013 & 17.5 & \textbf{9.0} & \textbf{18.4} & 0.013 & 19.8 \\
\bottomrule[1.5pt]
\end{tabular}
}
\caption{Performance across models and datasets by readers Llama-3.1-8B Instruct and Llama-3.3-70B Instruct with $k=5$. Compression Latency (Lat$\downarrow$) is in seconds, and Compression Ratio (Rate$\downarrow$) is reported as a percentage. Best results are in \textbf{bold}.}
\label{tab:k5}
\end{table*}

Tab.~\ref{tab:k5} reports the additional results under a smaller retrieval budget, $k=5$. Our method remains competitive across both readers, achieving the best results on several datasets and staying close to the strongest baseline in the remaining cases. It also keeps consistently low compression latency, second only to RECOMP-ext, while retaining a moderate portion of the context. These results suggest that the proposed pruner remains effective even when the initial retrieved context is limited.

Tab.~\ref{tab:gemini} provides additional results with stronger LLM readers, including Gemini-3-flash and GPT-5-mini under the low-effort setting. Across both datasets and readers, our method remains among the strongest compressors in answer quality while consistently achieving the lowest compression latency. AttnComp performs best with Gemini-3-flash, whereas our method is best or tied-best with GPT-5-mini. These results suggest that the proposed pruner remains effective with more capable readers and provides a favorable quality--efficiency trade-off.

\begin{table}[h]
\centering
\setlength{\tabcolsep}{4.5pt}
\resizebox{\linewidth}{!}{%
\begin{tabular}{l|cccc|cccc|c}
\toprule[1.5pt]
\multirow{2}{*}{Model} & \multicolumn{4}{c|}{HQA} & \multicolumn{4}{c|}{2Wiki} & \multirow{2}{*}{\begin{sideways}Reader\end{sideways}} \\
\cmidrule(lr){2-5} \cmidrule(lr){6-9}
& EM & F1 & Lat & Rate & EM & F1 & Lat & Rate & \\
\midrule
CompAct & 43.8 & 57.3 & 4.696 & \uline{3.7} & 45.9 & 51.7 & 4.905 & \uline{3.5} & \multirow{8}{*}{\begin{sideways}Gemini-3-flash (low)\end{sideways}} \\
EXIT & 42.2 & 56.0 & 1.465 & 34.0 & 42.4 & 49.0 & 1.492 & 29.4 &  \\
RECOMP-abs & 41.4 & 55.5 & 1.206 & \textbf{1.8} & 45.5 & 52.3 & 1.145 & \textbf{1.4} &  \\
Refiner & 43.9 & 57.8 & 3.915 & 4.4 & 47.8 & 54.5 & 3.342 & 3.7 &  \\
LongLLMLin & 42.3 & 56.6 & 1.104 & 30.0 & 42.7 & 49.8 & 1.133 & 30.2 &  \\
Provence & 44.2 & 58.4 & 0.199 & 13.8 & 47.6 & 54.1 & 0.197 & 11.9 &  \\
AttnComp & \textbf{47.4} & \textbf{62.4} & \uline{0.124} & 11.0 & \textbf{55.2} & \textbf{63.8} & \uline{0.130} & 7.0 &  \\
Ours & \uline{46.8} & \uline{61.7} & \textbf{0.045} & 13.8 & \uline{52.3} & \uline{59.0} & \textbf{0.046} & 11.5 &  \\
\midrule
CompAct & 41.8 & 57.1 & 4.696 & \uline{3.7} & 34.4 & 43.3 & 4.905 & \uline{3.5} & \multirow{8}{*}{\begin{sideways}GPT-5-mini (low)\end{sideways}} \\
EXIT & 40.3 & 55.8 & 1.465 & 34.0 & 36.7 & 47.5 & 1.492 & 29.4 &  \\
RECOMP-abs & 38.5 & 53.1 & 1.206 & \textbf{1.8} & 35.8 & 43.3 & 1.145 & \textbf{1.4} &  \\
Refiner & 39.9 & 54.8 & 3.915 & 4.4 & 35.8 & 44.0 & 3.342 & 3.7 &  \\
LongLLMLin & 40.1 & 55.5 & 1.104 & 30.0 & 37.1 & 48.6 & 1.133 & 30.2 &  \\
Provence & 43.0 & 58.9 & 0.199 & 13.8 & 41.9 & 50.9 & 0.197 & 11.9 &  \\
AttnComp & \uline{43.7} & \uline{60.1} & \uline{0.124} & 11.0 & \textbf{44.6} & \uline{54.9} & \uline{0.130} & 7.0 &  \\
Ours & \textbf{44.2} & \textbf{60.8} & \textbf{0.045} & 13.8 & \uline{44.6} & \textbf{54.9} & \textbf{0.046} & 11.5 &  \\
\bottomrule[1.5pt]
\end{tabular}
}
\caption{Performance across compressors on HotpotQA and 2WikiMultiHop datasets using Gemini-3-flash (low effort) and GPT-5-mini (low effort) readers with $k=20$.}
\label{tab:gemini}
\end{table}

\begin{table}[h]
    \centering
    \small
    \resizebox{\linewidth}{!}{
    \begin{tabular}{lcccc c}
        \toprule[1.5pt]
        Method & Memory (GB) $\downarrow$ & EM $\uparrow$ & F1 $\uparrow$ & Lat $\downarrow$ & Rate $\downarrow$ \\
        \midrule
        CompAct     & 14.34 & 34.23 & 43.85 & 4.777 & \underline{3.7} \\
        EXIT        & 11.53 & 34.09 & 44.32 & 1.451 & 35.1 \\
        RECOMP-abs  &  6.78 & 32.38 & 42.12 & 1.260 & \textbf{1.8} \\
        RECOMP-ext  & \textbf{0.88} & 29.76 & 39.17 & \textbf{0.036} & 11.9 \\
        Refiner     & 18.10 & 32.86 & 42.61 & 4.710 &  5.7 \\
        LongLLMLing & 14.48 & 35.06 & 45.55 & 1.113 & 30.1 \\
        Provence    &  1.95 & 34.73 & 45.08 & 0.194 & 15.5 \\
        AttnComp    &  6.80 & \textbf{37.16} & \textbf{47.65} & 0.125 &  9.3 \\
        Ours        & \underline{1.86} & \underline{36.98} & \underline{47.53} & \underline{0.045} & 16.0 \\
        \bottomrule[1.5pt]
    \end{tabular}
    }
    \caption{Detailed quality--latency--memory results corresponding to Fig.~\ref{fig:tradeoff}. EM, F1, and latency are averaged across five datasets at $k=20$ by Llama-3.3-70B-Instruct. Memory denotes the peak VRAM required. Rate denotes the compression rate (\%).}
    \label{tab:tradeoff_details}
\end{table}

Tab.~\ref{tab:tradeoff_details} reports the numerical values used to produce Fig.~\ref{fig:tradeoff} by Llama-3.3-70B-Instruct as LLM reader. The results show that our method attains competitive EM while requiring substantially lower peak memory ($\leq 2GB$) and compression latency ($\sim 45ms$) than most high-accuracy baselines, supporting its suitability for resource-constrained deployment. Similarly, we also provide results at $k=20$ with Llama-3.1-8B-Instruct as the reader in Tab.~\ref{tab:tradeoff_details_8b}.

\begin{table}[h]
    \centering
    \small
    \resizebox{\linewidth}{!}{
    \begin{tabular}{lcccc c}
        \toprule[1.5pt]
        Method & Memory (GB) $\downarrow$ & EM $\uparrow$ & F1 $\uparrow$ & Lat $\downarrow$ & Rate $\downarrow$ \\
        \midrule
        CompAct     & 14.34 & 30.29 & 39.43 & 4.777 & \underline{3.7} \\
        EXIT        & 11.53 & 30.05 & 38.79 & 1.451 & 35.1 \\
        RECOMP-abs  &  6.78 & 28.98 & 38.00 & 1.260 & \textbf{1.8} \\
        RECOMP-ext  & \textbf{0.88} & 25.75 & 33.62 & \textbf{0.036} & 11.9 \\
        Refiner     & 18.10 & 29.30 & 37.82 & 4.710 &  5.7 \\
        LongLLMLing & 14.48 & 30.78 & 39.51 & 1.113 & 30.1 \\
        Provence    &  1.95 & 30.86 & 39.65 & 0.194 & 15.5 \\
        AttnComp    &  6.80 & \underline{31.39} & \underline{40.27} & 0.125 &  9.3 \\
        Ours        & \underline{1.86} & \textbf{32.81} & \textbf{42.16} & \underline{0.045} & 16.0 \\
        \bottomrule[1.5pt]
    \end{tabular}
    }
    \caption{Detailed quality--latency--memory results with EM, F1, and latency are averaged across five datasets at $k=20$ by Llama-3.1-8B-Instruct. Memory denotes the peak VRAM required. Rate denotes the compression rate (\%).}
    \label{tab:tradeoff_details_8b}
\end{table}

The \textbf{codebase} and \textbf{model checkpoints} will be publicly released at \url{https://github.com/thaodod/EnComp} for replication.

\end{document}